\pdfoutput=1

\documentclass[11pt]{article}

\usepackage[preprint]{acl}

\usepackage{times}
\usepackage{latexsym}

\usepackage[T1]{fontenc}

\usepackage[utf8]{inputenc}

\usepackage{microtype}

\usepackage{inconsolata}

\usepackage{amsmath}
\usepackage{times}
\usepackage{framed}
\usepackage{ragged2e}
\usepackage{bm}
\usepackage{soul}
\usepackage{latexsym}
\usepackage{subfigure}
\usepackage{amsthm}
\usepackage{graphicx}
\usepackage{float}
\usepackage{longtable}
\usepackage{booktabs} 
\usepackage{multirow}
\usepackage{multicol}
\usepackage{lipsum}
\usepackage{amssymb}
\usepackage{dashbox}%
\usepackage{multirow}
\usepackage{textcomp}
\usepackage{tcolorbox}
\usepackage{bbding}
\usepackage{pifont}
\usepackage[ruled, vlined, linesnumbered]{algorithm2e}
\usepackage{mathtools}
\usepackage{adjustbox}
\usepackage[ruled,vlined]{algorithm2e}
\usepackage{color, soul}
\usepackage{array}
\usepackage{graphicx}
\usepackage{cleveref}
\usepackage{colortbl}
\usepackage{tabularx}
\usepackage{tablefootnote}
\crefname{section}{§}{§§}
\Crefname{section}{§}{§§}
\crefname{figure}{Figure}{Figure}
\Crefname{figure}{Figure}{Figure}
\crefname{table}{Table}{Table}
\Crefname{table}{Table}{Table}
\definecolor{forestgreen}{HTML}{228B22}

\newcommand\maven{\textsc{Maven}\xspace}
\newcommand\ourdata{\textsc{Maven-Fact}\xspace}
\newcommand\mavened{\textsc{Maven-ED}\xspace}
\newcommand\mavenere{\textsc{Maven-ERE}\xspace}
\newcommand\mavenarg{\textsc{Maven-Arg}\xspace}

\definecolor{my_red}{RGB}{255,99,71}
\definecolor{my_green}{RGB}{50,205,50}
\definecolor{my_blue}{RGB}{65,105,225}
\definecolor{ourdarkgreen}{RGB}{84,130,53}
\definecolor{ourdarkblue}{RGB}{68,114,195}

\title{\ourdata: A Large-scale Event Factuality Detection Dataset}

\author{Chunyang Li\thanks{\quad Equal contribution.}, Hao Peng$^{*}$, Xiaozhi Wang, Yunjia Qi, \\  \textbf{Lei Hou}, \textbf{Bin Xu}, \textbf{Juanzi Li}\\
Tsinghua University, Beijing, China \\
\texttt{\{lichunya20, peng-h21\}@mails.tsinghua.edu.cn}}

\begin{document}
\maketitle
\begin{abstract}

Event Factuality Detection (EFD) task determines the factuality of textual events, i.e., classifying whether an event is a fact, possibility, or impossibility, which is essential for faithfully understanding and utilizing event knowledge.
However, due to the lack of high-quality large-scale data, event factuality detection is under-explored in event understanding research, which limits the development of EFD community. 
To address these issues and provide faithful event understanding, we introduce \ourdata, a large-scale and high-quality EFD dataset based on the \maven dataset. \ourdata includes factuality annotations of $112,276$ events, making it the largest EFD dataset. Extensive experiments demonstrate that \ourdata is challenging for both conventional fine-tuned models and large language models (LLMs). Thanks to the comprehensive annotations of event arguments and relations in \maven, \ourdata also supports some further analyses and we find that adopting event arguments and relations helps in event factuality detection for fine-tuned models but does not benefit LLMs. 
Furthermore, we preliminarily study an application case of event factuality detection and find it helps in mitigating event-related hallucination in LLMs. Our dataset and codes can be obtained from \url{https://github.com/THU-KEG/MAVEN-FACT}

\end{abstract}

\section{Introduction}
\begin{figure}
    \centering
    \includegraphics[width=1.0\linewidth]{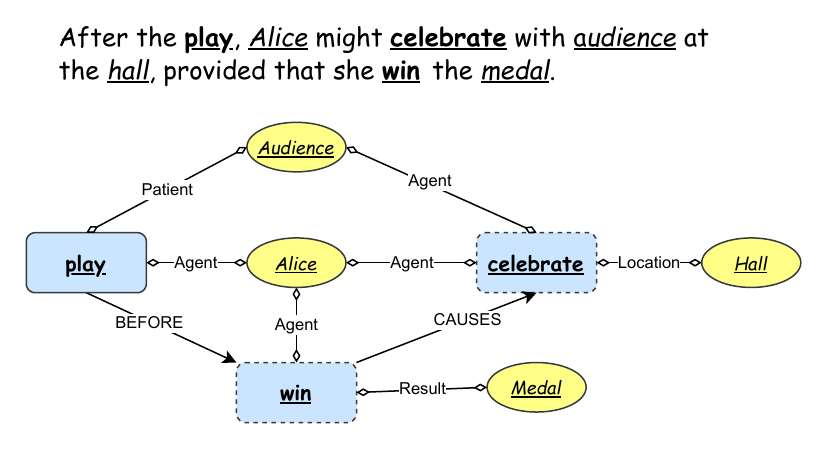}
    \caption{An example of event understanding. The event ``\textit{play}'' is factual while the events ``\textit{win}'' and ``\textit{celebrate}'' are just possibilities considering the word ``\textit{might}''.}
    \label{fig:figure1}
\end{figure}

Event Factuality Detection (EFD) aims to determine the factuality of textual events, i.e., classifying whether an event is a fact, possibility, or impossibility ~\citep{sauri2009factbank, sauri2012you, lee2015event, veyseh2019graph, murzaku2023towards}. As shown in Figure~\ref{fig:figure1}, the event ``\textit{play}'' is a fact while the event ``\textit{celebrate}'' is just a possibility considering the word ``\textit{might}''.

Event factuality detection is a subfield of event understanding, which aims to extract structured event knowledge from plain texts~\citep{wang2023maven, wang2023code4struct, peng2023devil, huang2023reevaluation, choudhary2024qaevent}, as shown in Figure~\ref{fig:figure1}. Event understanding is fundamental to broad downstream applications~\citep{ding2015deep, goldfarb-tarrant-etal-2019-plan, wang2021incorporating}. 
Previous event understanding work focuses on three primary tasks: event detection~\citep{wang2020maven}, event argument extraction~\citep{wang2023maven}, and event relation extraction~\citep{wang2022maven}.
However, event factuality detection is under-explored. 

The primary reason for the under-exploration of EFD may be the lack of a large-scale, high-quality EFD dataset. Previous EFD datasets are usually small-scale. For example, the most widely-used dataset FactBank~\citep{sauri2009factbank} only includes $9,761$ events, which may not provide sufficient data for model training and evaluation. 
Furthermore, these datasets also lack annotations for event arguments and relations, preventing a comprehensive understanding of events. In fact, considering factuality is crucial in event understanding. For example, if a downstream application mistakenly takes the ``\textit{celebrate}'' event in Figure~\ref{fig:figure1} as a fact rather than a possibility, it is likely to lead to erroneous judgments or even broader impacts.

To alleviate these issues, we introduce \ourdata, a large-scale and high-quality event factuality detection dataset based on \maven~\citep{wang2020maven, wang2022maven, wang2023maven}. \maven provides a unified and comprehensive annotation, including event types~\citep{wang2020maven}, arguments~\citep{wang2023maven}, and relations~\citep{wang2022maven}, for the same set of documents. Building on the solid foundation of the \maven series, this work extends the annotation to include event factuality. Therefore, \ourdata includes comprehensive annotations including event types, arguments, relations, and factuality, which can support faithful all-in-one event understanding. \ourdata also offers three main advantages:
(1) \textbf{Large data scale}. \ourdata includes factuality annotations for $112,276$ events, making it the largest EFD dataset. 
(2) \textbf{Supporting evidence annotation}.
\ourdata also provides supporting evidence annotations, i.e., the words that directly convey factuality, e.g., \textit{may}, for non-factual events. 
This enables 
a detailed analysis of factuality understanding of models and enhances models' interpretability by outputting supporting evidence of their factuality predictions~\citep{zhao2024explainability}.
(3) \textbf{Enabling task interaction}. 
Intuitively, some event information may help in factuality detection. For example, if an event has the \textit{start-time} argument, then the event should be a fact. 
Thanks to \maven's event annotations, \ourdata enables analyzing how the event elements, including type, arguments, and relations, affect factuality detection, and vice versa.

To reduce cost and ensure data quality, we design an LLM-then-human annotation approach. Specifically, due to most events (exceeds $80\%$) being factual, we can endeavor to pre-annotate them automatically. We employ GPT-3.5~\citep{chatgpt} for pre-annotation and formalize the task as a binary classification task (\texttt{factual} or \texttt{non-factual}) and develop a chain-of-thought prompt~\citep{wei2022chain} method incorporating heuristic rules to ensure the high recall rate of the \texttt{non-factual} class. Subsequently, we manually annotate events pre-annotated as \texttt{non-factual}. To ensure data quality, the LLM pre-annotation is only used for the training set, while the events in validation and test sets are all human-annotated. This approach saves about $15\%$ annotation costs (about $2,500$ USD).

In the experiments, we evaluate several strong and representative models, including fine-tuned EFD models~\citep{kenton2019bert, liu2019roberta, wang-etal-2019-hmeae, murzaku2023towards}, and LLMs with in-context learning~\citep{brown2020language}, including Mistral 7B~\citep{jiang2023mistral}, LLAMA 3~\citep{llama-3}, GPT-3.5~\citep{chatgpt}, and GPT-4~\citep{openai2023gpt}. Experimental results demonstrate that the best-performing model only achieves a $47.6\%$ macro F1 score and an even lower F1 score for non-factual events.
It suggests that \ourdata is quite challenging for existing EFD models and LLMs. We conduct further experiments by requiring the models 
to provide supporting words for their predictions and find that this F1 score is much lower than that of EFD.
It indicates that even if the model can correctly detect factuality, it may not provide accurate explanations. 
We also observe that adding arguments and relations enhances the performance of fine-tuned EFD models, whereas it does not benefit LLMs with in-context learning.
Furthermore, we preliminarily study a potential application case of event factuality detection in mitigating event-related hallucination~\citep{huang2023survey}, and find that incorporating event factuality can help mitigate hallucination in LLMs.
We hope \ourdata and our empirical findings could facilitate future research on event factuality detection and faithful event understanding.

\section{Dataset Construction}


This section introduces the definition of event factuality detection (\cref{sec:task_form}), the LLM-then-human annotation approach (\Cref{sec:llm_then_human,sec:llm_ann,sec:human_ann}), and data analysis of \ourdata (\cref{sec:data_analysis}).

\subsection{Task Formulation}
\label{sec:task_form}
\begin{figure}
    \centering
    \includegraphics[width=0.5\linewidth]{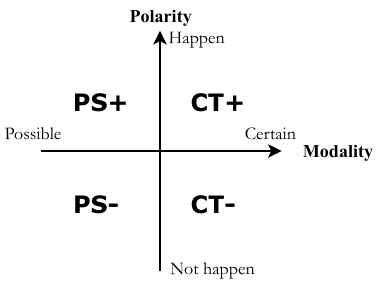}
    \caption{An illustration of four factuality classes. \textit{Uu} denotes factuality can not be determined by the given context and is not shown in the figure.}
    \label{fig:fact_class}
\end{figure}

Event factuality detection is the task of assessing whether an event is a fact. Typically, this task is formalized as a multi-class classification problem. We adopt the widely-used $5$ classes~\citep{sauri2009factbank, qian2018event}, \textit{CT+}, \textit{PS+}, \textit{PS-}, \textit{CT-}, and \textit{Uu}, as the label set of \ourdata.
These classes are based on the polarity and modality of event factuality, as illustrated in Figure~\ref{fig:fact_class}.  To reduce annotation cost and bias, we do not adopt finer-grained class definitions like in \citet{lee2015event}. These $5$ classes are sufficient to express the polarity and modality for factuality detection and support its applications. \ourdata also supports the supporting evidence prediction task~\citep{alvarez2018towards}, which predicts the words conveying the factuality. In this paper, we formalize this task as a pipeline, where models first perform EFD and then predict supporting words based on their factuality predictions.

\subsection{LLM-then-Human Annotation Approach}
\label{sec:llm_then_human}
In this paper, we aim to annotate the factuality of the overall $112,276$ events from the \maven dataset to construct \ourdata. Due to the large scale of the data, manual annotation for all events is costly and not conveniently transferable to other domains or scenarios. 
Given the proven efficacy of LLMs as effective annotators~\citep{mirzakhmedova2024large, chen2024large}, we develop an LLM-then-human annotation workflow to reduce costs while ensuring the annotation quality. We first adopt GPT-3.5~\citep{chatgpt} to pre-annotate the data, filtering out events requiring human annotation, followed by a meticulous human annotation. This annotation approach reduces annotation costs by approximately $15\%$, saving about $2,500$ USD. We only annotate the supporting words for \textit{PS+}, \textit{PS-}, and \textit{CT-} events, as \textit{CT+} and \textit{Uu} events usually do not involve obvious supporting evidence.
We employ this annotation workflow for the training set events only, while the validation and test sets are fully human-annotated. We finally sample $50$ documents from the training set and find less than $2\%$ noise, which demonstrates the efficacy of our annotation approach and the high quality of \ourdata. We will describe the details of LLM annotation (\cref{sec:llm_ann}) and human annotation (\cref{sec:human_ann}) in the following sections. 

\begin{figure}
    \centering
\includegraphics[width=0.75\linewidth]{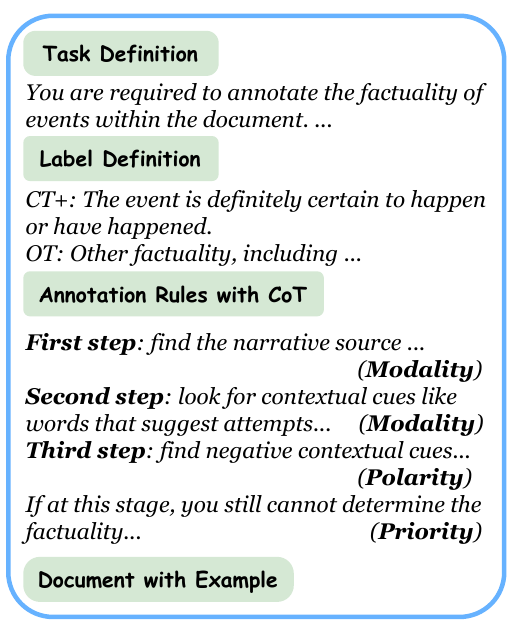}
    \caption{Prompt used in LLM pre-annotation.}
    \label{fig:cot_prompt}
\end{figure}

\subsection{LLM Annotation}
\label{sec:llm_ann}

Due to the majority of events being \textit{CT+}, i.e., already occurred, we aim to employ LLMs to \textbf{automatically} distinguish between \textit{CT+} and \textit{non-CT+} events, and \textbf{only events pre-annotated as non-CT+ require further human annotation}, thereby reducing the need for human annotation. Consequently, the recall of pre-annotating \textit{non-CT+} events is crucial for reducing pre-annotation noise. We improve the recall of \textit{non-CT+} events in two main aspects: (1) We simplify the event factuality detection task into a binary classification problem, only distinguishing between \textit{CT+} and \textit{non-CT+} factuality, as binary classification is generally simpler than multi-classification~\citep{rifkin2004defense}. Our objective is to obtain a high recall score for \textit{non-CT+} to avoid filtering events that require further human annotation. 
(2) We adopt the chain-of-thought prompting method (CoT)~\citep{wei2022chain} to better promote LLMs. Specifically, we first construct a comprehensive collection of annotation rules for the EFD task, integrated from multiple authors. Based on these rules, we design a step-by-step prompt that mirrors the human process for this task: {\normalsize{\textcircled{\scriptsize{\textbf{1}}}}} Determine the narrative perspective of the event.
{\normalsize{\textcircled{\scriptsize{\textbf{2}}}}}
Identify words conveying modalities, such as ``may''.
{\normalsize{\textcircled{\scriptsize{\textbf{3}}}}}
Check whether the text contains negations.
{\normalsize{\textcircled{\scriptsize{\textbf{4}}}}}
Default classification to \textit{non-CT+} if factuality cannot be determined. 
Figure~\ref{fig:cot_prompt} illustrates the details of the prompt.

\begin{table}
    \centering
    \small
    \begin{tabular}{l|cc|cc}
    \toprule
    \multirow{2}{*}{\textbf{Setting}} & \multicolumn{2}{c|}{\textbf{Direct Prompt}}  & \multicolumn{2}{c}{\textbf{CoT Prompt}} \\
    \cmidrule{2-5}
    & non-CT+ & CT+ & non-CT+ & CT+ \\
    \midrule
    Multi-Class & $18.8$ & $95.7$ & $55.2$ & $69.2$ \\
    Bi-Class & $27.1$ & $89.3$ & $97.4$ & $16.5$ \\
    \bottomrule
    \end{tabular}
    \caption{The recall rate of \textit{non-CT+} and \textit{CT+} using direct and CoT prompts under multiple and binary classification settings. Higher \textit{non-CT+} recall denotes less pre-annotation noise. Higher \textit{CT+} recall means reducing more human annotation.}
    \label{tab:prelimi_study}
\end{table}

\begin{table*}[t]
    \centering
    \small
                \begin{adjustbox}{max width=1\linewidth}
{
    \begin{tabular}{l|r|rrrrr|r|c}
    \toprule
    \textbf{Dataset} & \textbf{\#Doc.}& \textbf{\#CT+} & \textbf{\#CT-} & \textbf{\#PS+} & \textbf{\#PS-} & \textbf{\#Uu} & \textbf{\#Total} & \textbf{Supporting Words}  \\
    \midrule
    FactBank & $208$& $7,749$ & $433$ & $589$ & $70$ & $4,619$ & $13,460$ & \ding{55}  \\
    MEANTIME & $120$ & $1,798$ & $44$ & $83$ & $3$ & $125$ & $2,053$ & \ding{55} \\
    UW & $276$ & $-$ & $-$ & $-$ & $-$ & $-$ & $13,923$ &  \ding{55}  \\
    UDS-IH2 & $-$ & $-$ & $-$ & $-$ & $-$ & $-$ & $27,289$ & \ding{55} \\
    EB-DLEF & $7,840$ & $5,222$ & $1,601$ & $935$ & $53$ & $29$ & $7,840$ &  \ding{51}  \\
    DLEF-v2 & $9,180$ & $5,555$ & $2,029$ & $1,454$ & $84$ & $58$ & $9,180$ & 
    \ding{51}  \\
    \midrule
    \ourdata & $4,480$ & $105,209$ & $2,330$ & $3,874$ & $540$ & $323$ & $112,276$ & \ding{51}  \\
    \bottomrule
    \end{tabular}
}
    \end{adjustbox}
    \caption{Statistics of \ourdata compared with other event factuality detection datasets. Doc. is the short for Document. ``Supporting Words'' means whether the dataset contains supporting words of factuality. UW and UDS-IH2 adopt continuous factuality values and hence the statistics for discrete factuality are not applicable.}
    \label{tab:statistics}
\end{table*}

To validate the efficacy of our annotation method, we conduct a pilot experiment on $500$ human-annotated events. We adopt GPT-3.5 as the LLM for pre-annotating. The experimental results are presented in Table~\ref{tab:prelimi_study}. We can observe that our approach, binary classification with chain-of-thought prompting, achieves a $97.5\%$ recall of \textit{non-CT+}, which indicates it introduces little noise during pre-annotation, and a $16.5\%$ recall of \textit{CT+}, which suggests that the pre-annotation substantially reduce the annotation cost. Finally, we pre-annotate the overall training set, resulting in $16,950$ events labeled as \textit{CT+} and $56,989$ as \textit{non-CT+}, with the latter needing further human annotation.


\subsection{Human Annotation}
\label{sec:human_ann}
To ensure data quality, we manually annotate all events pre-labeled as \textit{non-CT+} in the training set and all events in the validation and test sets. We employ a commercial annotation company for annotation, which involves $47$ annotators, including $8$ senior annotators responsible for annotation training and quality verification of other annotators's annotations. All annotations were performed on a specially developed platform. We ask the annotators to assess the factuality of events based solely on the provided text, without considering external knowledge. If annotators can not determine the factuality based on the provided text, they will label the event as \textit{Uu}.
For \textit{PS+}, \textit{PS-}, and \textit{CT-} events, annotators are required to provide supporting evidence, which are the words extracted from the given text. If there are multiple supporting words, annotators are required to extract all of them. Senior annotators will randomly check $5\%$ of annotated factuality. If an annotator’s error rate assessed by the senior annotator exceeds $5\%$, they will undergo re-training for the annotation and all of this annotator's annotation will be re-labeled. We randomly sample $100$ documents and annotate them twice by different annotator groups. The final inter-annotator agreement (accuracy) is $96.1\%$, demonstrating the annotation quality. Annotation details are provided in \cref{sec:app_data}.

\subsection{Data Analysis}
\label{sec:data_analysis}

Table~\ref{tab:statistics} presents the statistics of \ourdata and other widely-used event factuality detection datasets, including FactBank~\citep{sauri2009factbank}, MEANTIME~\citep{minard2016meantime}, UW~\citep{lee2015event}, EB-DLEF~\citep{zhang2022evidence}, DLFE-v2~\citep{qian2022document}, UDS-IH2~\citep{rudinger2018neural}. We can observe that \ourdata possesses the largest data scale and includes supporting word annotations. Thanks to \maven's extensive annotations, \ourdata provides comprehensive annotations of events, arguments, relations, and factuality, supporting comprehensive and faithful event understanding research and applications.

\section{Experiment}

\begin{table*}[t]
\small
\centering
\begin{adjustbox}{max width=1\linewidth}{
\begin{tabular}{l|rrr|rrr|rrr|rrr|rrr|c}
\toprule
 \multirow{2}{*}{\textbf{Model}} & \multicolumn{3}{c|}{\textbf{CT+}} & \multicolumn{3}{c|}{\textbf{CT-}} & \multicolumn{3}{c|}{\textbf{PS+}} & \multicolumn{3}{c|}{\textbf{PS-}} & \multicolumn{3}{c|}{\textbf{Uu}} & \multirow{2}{*}{\textbf{Macro-F1}} \\
\cmidrule(l){2-16}
 & P & R & F1 & P & R & F1 & P & R & F1 & P & R & F1 & P & R & F1 & \\ 
\midrule
    BERT+CLS  & $94.1$ & $98.6$ & $96.3$ & $\mathbf{66.6}$ & $54.0$ & $59.6$ & $61.4$ & $35.5$ & $45.0$ & $\mathbf{61.0}$ & $17.7$ & $27.5$ & $15.4$ & $1.1$ & $2.0$ & $46.1$ \\
    RoBERTa+CLS  & $94.1$ & $98.6$ & $96.3$ & $61.0$ & $54.8$ & $57.8$ & $61.3$ & $32.0$ & $42.0$ & $44.8$ & $19.2$ & $26.9$ & $\mathbf{50.0}$ & $2.2$ & $4.1$ & $45.4$ \\
    DMBERT  & $94.4$ & $98.4$ & $96.3$ & $64.8$ & $55.9$ & $60.0$ & $62.2$ & $37.5$ & $46.8$ & $45.6$ & $23.2$ & $30.7$ & $26.7$ & $2.2$ & $3.1$ & $\mathbf{47.6}$ \\
    DMRoBERTa  & $94.3$ & $98.4$ & $96.3$ & $62.3$ & $\mathbf{60.4}$ & $\mathbf{61.3}$ & $62.6$ & $34.4$ & $44.4$ & $50.0$ & $23.2$ & $\mathbf{31.6}$ & $16.7$ & $1.1$ & $2.0$ & $47.1$ \\
    GenEFD  & $94.2$ & $\mathbf{98.7}$ & $\mathbf{96.4}$ & $65.9$ & $54.4$ & $59.6$ & $\mathbf{63.8}$ & $37.5$ & $47.3$ & $57.1$ & $13.8$ & $22.2$ & $0.0$ & $0.0$ & $0.0$ & $45.1$ \\
\midrule
    Mistral 7B  & $92.6$ & $80.7$ & $86.2$ & $30.0$ & $11.8$ & $16.9$ & $14.4$ & $39.8$ & $21.2$ & $3.5$ & $17.6$ & $5.8$ & $14.3$ & $6.1$ & $\mathbf{8.5}$ & $27.7$ \\
    \quad \hspace{10pt}\texttt{+CoT} & $93.0$ & $70.3$ & $80.1$ & $9.0$ & $17.6$ & $11.9$ & $11.6$ & $38.3$ & $17.8$ & $2.6$ & $23.5$ & $4.8$ & $10.5$ & $6.1$ & $7.7$ & $24.4$ \\
    LLAMA 3 & $91.7$ & $75.3$ & $82.7$ & $18.2$ & $11.8$ & $14.3$ & $10.2$ & $38.3$ & $16.1$ & $0.0$ & $0.0$ & $0.0$ & $7.7$ & $6.1$ & $6.8$ & $24.0$ \\
    \quad \hspace{10pt}\texttt{+CoT} & $95.8$ & $62.6$ & $75.8$ & $25.0$ & $17.5$ & $20.7$ & $12.3$ & $\mathbf{71.9}$ & $21.0$ & $11.8$ & $23.5$ & $15.7$ & $1.5$ & $3.0$ & $2.0$ & $27.0$ \\
    GPT-3.5 & $94.1$ & $53.0$ & $67.8$ & $3.6$ & $54.9$ & $6.7$ & $12.5$ & $7.8$ & $9.6$ & $1.
    2$ & $11.8$ & $2.1$ & $3.7$ & $3.0$ & $3.3$ & $17.9$ \\
    \quad \hspace{10pt}\texttt{+CoT} & $\mathbf{96.1}$ & $27.9$ & $43.2$ & $4.1$ & $36.3$ & $7.3$ & $8.4$ & $47.7$ & $14.3$ & $3.1$ & $\mathbf{52.9}$ & $5.8$ & $3.3$ & $\mathbf{10.6}$ & $5.1$ & $15.1$ \\
    GPT-4  & $94.1$ & $94.6$ & $94.4$ & $51.4$ & $37.3$ & $43.2$ & $44.7$ & $56.3$ & $49.8$ & $16.7$ & $11.8$ & $13.8$ & $0.0$ & $0.0$ & $0.0$ & $40.2$ \\
    \quad \hspace{10pt}\texttt{+CoT} & $94.8$ & $94.2$ & $94.5$ & $46.5$ & $39.2$ & $42.6$ & $43.4$ & $58.6$ & $\mathbf{49.8}$ & $20.0$ & $23.5$ & $21.6$ & $25.0$ & $3.0$ & $5.4$ & $42.8$ \\
 
\bottomrule
\end{tabular}
}
\end{adjustbox}
\caption{Experimental results of fine-tuned EFD models and LLMs with in-context learning on \ourdata.}
\label{tab:main_res}
\end{table*}

\subsection{Experimental Setup}
\paragraph{Baselines}
We evaluate several advanced and representative models, mainly including fine-tuned EFD models and large language models with in-context learning~\citep{brown2020language}. 

For fine-tuned EFD models, we reproduce several advanced models, including (1) \textbf{BERT+CLS} and \textbf{RoBERTa+CLS}, which adopt BERT~\citep{kenton2019bert} and RoBERTa~\citep{liu2019roberta} as the text encoder, respectively, and use the representation of a special token \texttt{[CLS]}~\citep{kenton2019bert} as the factuality representation of the event for factuality classification.
(2) \textbf{DMBERT}~\citep{wang-etal-2019-hmeae} and \textbf{DMRoBERTa}, classical event understanding models that also utilize BERT and RoBERTa as the text encoder, respectively, and incorporate a dynamic multi-pooling mechanism~\citep{chen2015event} to integrate context and event information into a factuality representation for the final factuality classification.
(3) \textbf{GenEFD}~\citep{murzaku2023towards}, a generative model based on FLAN-T5~\citep{chung2024scaling}. \citet{murzaku2023towards} transform the event factuality detection task into a text generation task and design a meticulous factuality structure and target text structure, then optimize FLAN-T5 through multi-task learning. This model achieves state-of-the-art performance on FactBank. Our implementation employs the same setting except for not using multi-task learning as we only train the model on the event factuality detection task. 
For encoder-only models, we utilize cross-entropy loss for training. For training GenEFD, we employ language modeling loss~\citep{bengio2000neural}.

We also evaluate several LLMs with in-context learning, including two powerful open-sourced LLMs, \textbf{Mistral 7B}~\citep{jiang2023mistral} and the 8B version of \textbf{LLAMA 3}~\citep{llama-3}, and two proprietary LLMs, \textbf{GPT-3.5}~\citep{chatgpt} and \textbf{GPT-4}~\citep{openai2023gpt}. For all experiments, we adopt $5$-shot in-context learning. The demonstrations contain one exemplar from each category and are randomly sampled from the training set of \ourdata. We also evaluate LLMs with chain-of-thought prompt method~\citep{wei2022chain}, which is the same as in \cref{sec:llm_ann} used for data annotation. 
Considering the time and monetary costs for LLMs inference, we sample $2,000$ instances from the original test set of \ourdata to evaluate LLMs. More details of the experimental setup are placed in \cref{sec:app_efd}.

\paragraph{Evaluation Setup}
We adopt the same evaluation metrics with previous work~\citep{qian2018event}, and report precision (P), recall (R), F1 scores, and their macro averages for the \textit{CT+}, \textit{CT-}, \textit{PS+}, and \textit{PS-}, and \textit{Uu} categories. For generative models, we use the exact match method~\citep{rajpurkar-etal-2016-squad} to compute the consistency rate between outputs and ground truth labels. For the chain-of-thought outputs, we require the LLM to provide its answer directly after ``answer:'' and automatically parse its response. If the outputs do not conform to this format, they are categorized as \textit{Uu} predictions.

\subsection{Experimental Results}

The experimental results are shown in Table~\ref{tab:main_res}, and we have the following observations:

(1) Both the fine-tuned EFD models and LLMs exhibit moderate performance, particularly in the \textit{CT-}, \textit{PS+}, and \textit{PS-} categories, compared to the results in the widely-used FactBank dataset~\citep{qian2018event, murzaku2023towards}. This suggests that \ourdata poses a significant challenge to existing models. The small scale of existing datasets with limited \textit{non-CT+} data may be insufficient for training and benchmarking EFD models, and hinders the development of advanced models. We hope the large-scale \ourdata data will facilitate more research efforts to develop advanced models for the event factuality detection task.
(2) LLMs significantly underperform fine-tuned models, especially in the \textit{non-CT+} categories, and even the most powerful GPT-4 only achieves $42.8\%$ macro F1. This aligns with previous findings that LLMs with in-context learning often fall short in information extraction tasks~\citep{li2023evaluating, han2023information}, possibly because LLMs lack specific understanding abilities to fine-grained information~\citep{peng2023specification}, which is necessary for detecting factuality, such as ``may''. This suggests that LLMs may confuse event factuality, and we will show in \cref{sec:hallucination} that it results in non-factual responses of LLMs, i.e., hallucinations~\citep{huang2023survey}, into the tasks requiring event knowledge.
(3) The chain-of-thought approach significantly improves LLMs' performance. One possible reason is that the detailed instructions in the prompt enhance LLMs' fine-grained comprehension of texts, such as supporting words. Although these results are still well below those of fine-tuned models, it suggests that designing meticulous prompts to enhance the event factuality understanding ability of LLMs is feasible, and more research efforts are needed for enhancing this capability of LLMs, such as utilizing \ourdata as high-quality alignment data to align LLMs on the EFD task~\citep{qi2024adelie}.

In conclusion, \ourdata presents a significant challenge to existing EFD models and LLMs. We hope that the high-quality \ourdata dataset will contribute to the training and benchmarking of EFD models and call for more research efforts to develop advanced EFD models.

\subsection{Supporting Evidence Prediction}
There are numerous works exploring explainability for models by requiring the models to provide explanations, i.e., supporting evidence, for their outputs, thereby enhancing the interpretability,  transparency, and reliability of models~\citep{luo2024understanding}. It is particularly essential for tasks involving factuality-related outputs where models are prone to generating hallucinations~\citep{huang2023survey}.
Therefore, for event factuality detection, providing coherent supporting evidence is essential for assessing the inherent understanding of event factuality and improving the reliability of models.
However, as shown in Table~\ref{tab:statistics}, most previous datasets lack annotations for supporting evidence, i.e., the words conveying the factuality, leading to a lag in related research. \ourdata comprehensively provide annotated supporting words for \textit{CT-}, \textit{PS+}, and \textit{PS-} events to facilitate research on predicting supporting words and developing reliable EFD models. 

\begin{table}[t]
\centering
\small
\scalebox{0.8}{
\begin{tabular}{l|rrr|rrr}
\toprule
    \multirow{2}{*}{\textbf{Model}} & \multicolumn{3}{c|}{\textbf{Factuality}} & \multicolumn{3}{c}{\textbf{Supporting Evidence}} \\
    \cmidrule(l){2-7}
    & P & R & F1 & P & R & F1 \\
    \midrule
    DMRoBERTa  & $74.5$ & $\mathbf{49.1}$ & $\mathbf{58.1}$ & $\mathbf{55.8}$ & $39.4$ & $\mathbf{45.4}$ \\
    GenEFD  & $\mathbf{76.3}$ & $40.5$ & $50.4$ & $49.5$ & $\mathbf{40.8}$ & $44.7$ \\
    LLAMA 3  & $53.7$ & $14.3$ & $18.5$ & $4.6$ & $2.8$ & $3.5$ \\
    GPT-4  & $62.4$ & $32.6$ & $42.5$ & $21.0$ & $18.3$ & $19.5$ \\
    \bottomrule
\end{tabular}
}
\caption{Macro averages of precision (P), recall (R), and F1 scores of \textit{CT-}, \textit{PS+}, \textit{PS-} on the factuality and supporting evidence prediction task. 
We report the macro averages for only these three categories because only they have supporting evidence in the given input text.}
\label{tab:sw_res}
\end{table}

We evaluate the EFD models on the task of predicting supporting words for their factuality predictions in the \ourdata dataset. Specifically, we employ a pipeline form where the model first detects the event factuality and then predicts supporting words based on its predicted factuality. Given that an event may have multiple supporting words, we use the sequence labeling paradigm~\citep{akhundov2018sequence} for models to predict these words. Without loss of generality, we evaluate four representative models, including DMRoBERTa, GenEFD, LLAMA 3, and GPT-4.
Further experimental details are provided in \cref{sec:app_efd}. Table~\ref{tab:sw_res} presents the results, and we can find that the performance of supporting word prediction is significantly inferior to that of event factuality detection. This indicates that providing supporting words is more challenging, and models may struggle to provide valid supporting evidence even when they accurately predict factuality.

We further conduct an error analysis of supporting word prediction, and the errors stem from two main sources: incorrect factuality prediction and incorrect supporting word prediction. We categorize the errors into three types: \texttt{OnlyF}, \texttt{OnlyW}, and \texttt{Both}, which denote the errors come from only factuality prediction, only supporting word prediction, and both, respectively. The results are presented in Table 4. We can observe that (1) About $30\%$ of the errors are \texttt{OnlyW}, indicating that even if the model accurately predicts factuality, it may still struggle to correctly predict the supporting words. (2) Except for LLAMA 3, a significant portion of the errors are \texttt{OnlyF}, suggesting that although the model does not predict factuality correctly, it accurately identifies supporting words. These errors suggest that the models can not sufficiently explain their own outputs. This hurts the reliability of the model in event factuality prediction. 
More efforts are needed to develop more reliable EFD models.

\begin{table}[t]
\centering
\small
\begin{tabular}{l|rrr}
\toprule
\textbf{Model} & \textbf{\texttt{OnlyF}} & \textbf{\texttt{OnlyW}} & \textbf{\texttt{Both}} \\
\midrule
DMRoBERTa & $6.8$ & $31.0$ & $62.2$ \\
GenEFD & $24.6$ & $31.6$ & $43.8$ \\
LLAMA 3 & $0.6$ & $25.4$ & $74.0$ \\
GPT-4 & $9.6$ & $37.6$ & $52.9$ \\
\bottomrule
\end{tabular}
\caption{Error rate (\%) on supporting word prediction. \texttt{OnlyF}, \texttt{OnlyW}, and \texttt{Both} mean the errors come from only factuality prediction, only supporting word prediction, and both, respectively.}
\label{tab:error_sw}
\end{table}

\subsection{Analysis on Task Interaction}
Thanks to \maven's comprehensive annotations, \ourdata also facilitates research about the interactions between event elements, such as arguments and relations, and event factuality, which is under-explored previously due to a lack of comprehensive data. In this paper, we primarily investigate whether event arguments and relations can help in event factuality detection. Intuitively, additional event information can benefit EFD. For example, if an event has a ``\textit{time}'' argument, it suggests the event is a fact that has already happened.

\begin{table}[t]
\centering
\small
\begin{tabular}{l|rrr}
\toprule
\textbf{Model} & \textbf{Precision} & \textbf{Recall} & \textbf{Macro F1} \\
\midrule
DMRoBERTa & $57.2$ & $43.5$ & $47.1$ \\
\quad \texttt{+relation} & $63.7$ & $44.1$ & $49.1$ \\
\quad \texttt{+argument} & $63.2$ & $45.4$ & $49.3$ \\
\quad \texttt{+both} & $53.5$ & $41.2$ & $45.6$ \\
\midrule
GenEFD & $56.2$ & $40.9$ & $45.1$ \\
\quad \texttt{+relation} & $57.7$ & $41.4$ & $45.4$ \\
\quad \texttt{+argument} & $54.4$ & $43.0$ & $46.4$ \\
\quad \texttt{+both} & $53.8$ & $44.7$ & $47.6$ \\
\midrule
LLAMA 3 & $25.6$ & $26.3$ & $24.0$ \\
\quad \texttt{+relation} & $21.9$ & $25.3$ & $11.6$ \\
\quad \texttt{+argument}  & $23.1$ & $29.4$ & $19.4$ \\
\quad \texttt{+both} & $23.5$ & $25.1$ & $16.9$ \\
\midrule
GPT-4 & $41.4$ & $40.0$ & $40.2$ \\
\quad \texttt{+relation} & $42.4$ & $35.9$ & $37.9$  \\ 
\quad \texttt{+argument} & $41.4$ & $36.0$ & $37.6$ \\
\quad \texttt{+both} & $43.2$ & $37.6$ & $39.7$ \\
\bottomrule
\end{tabular}
\caption{Performance on event factuality detection after adding different event information.}
\label{tab:tl_res}
\end{table}

We conduct experiments on four representative models, DMRoBERTa, GenEFD, LLAMA 3, and GPT-4, using \ourdata to investigate whether adding event arguments and relations can help in EFD. For GenEFD, LLAMA 3 and GPT-4, we introduce additional information by transforming arguments and relations into natural language forms and placing them in the original text input.
For DMRoBERTa, except for adding them in the text input, 
we introduce additional information by concatenating the representations of arguments and relations to the event factuality representation, and then use this concatenated representation to classify the factuality. The representations of arguments and relations are the average representations of their corresponding tokens. More experimental details can be found in \cref{sec:app_efd}.

The results are shown in Table~\ref{tab:tl_res}, and we have the following observations: 
(1) For fine-tuned EFD models, DMRoBERTa and GenEFD, the experimental results generally align with our expectations, where the introduction of event arguments or relations tends to boost factuality detection performance. It suggests that fine-tuning models could better learn these correlations. However, adding both relation and argument information hurts the performance of DMRoBERTa. One possible reason is that the concatenated relation and argument representations may cause the model to more easily overfit to certain patterns in the training set. (2) For LLMs with in-context learning, introducing additional information tends to worsen performance. We find that the decline primarily comes from \textit{CT+}, and the models are sensitive to prompts~\citep{dong2022survey} and shifted towards classifying factuality as \textit{non-CT+}. This suggests that few-shot in-context learning might introduce some biases instead of generalizable patterns~\citep{si2023measuring}. More efforts are needed to effectively introduce additional information by in-context learning for event factuality detection, such as using many-shot in-context learning~\citep{agarwal2024many}.

\section{Mitigating Event-related Hallucinations}
\label{sec:hallucination}
In addition to benchmarking EFD models, we also want to explore potential application scenarios of \ourdata. 
Here, we preliminarily explore using event factuality to mitigate event-related hallucinations in LLMs, as non-factuality is a primary source of hallucination~\citep{huang2023survey}.

Hallucination refers to the phenomenon where the outputs of models do not align with the input, typically involving non-factual information in the outputs~\citep{huang2023survey}. This issue is prevalent in existing LLMs, raising concerns about the reliability and faithfulness of LLMs. 
There are numerous works exploring detecting and mitigating hallucination in LLMs~\citep{ji2023towards,dhuliawala2023chain,yang2023alignment,zhang2024r,li2024inference}. Event-related hallucination refers to the model outputting incorrect information about an event, such as erroneous event arguments or causal relations, which is under-explored in previous research.
Here, we hope to explore whether providing explicit event factuality information can help mitigate event-related hallucinations in LLMs.


\begin{table}
    \centering
    \small
    \begin{tabularx}{\linewidth}{X}
        \toprule
             \textbf{Document:}
             The 2014 Bukidnon bus bombing occurred on December 9, 2014. ... Extortion is viewed as a motive for the attacks due to 
            \textit{\color{blue}claims} that the bus company has faced threats for refusing to \textit{\color{blue}pay} protection money to the militants. The militant group denies any involvement claiming they would not gain any benefit from conducting such attacks and claims the accusations against them as fabrication. \\
        \midrule
            \textbf{\color{ourdarkgreen}Question:}
            Did the bus bombing occur because the bus company refused to \textit{\color{blue}pay} protection money to the militants? \\
        \midrule \textbf{\color{ourdarkblue}Answer:}
            No \\
        \bottomrule 
    \end{tabularx}
    \caption{An instance from our constructed QA dataset. The ``\textit{\color{blue}pay}'' event is a \textit{PS-} event. 
    }
    \label{tab:example_qa}
\end{table}

\paragraph{Experimental Setup}
We begin with constructing a knowledge-intensive question-answering (QA) dataset based on \ourdata, which is a scenario susceptible to hallucinations~\citep{huang2023survey}. 
As we aim to analyze hallucination in LLMs, we deliberately craft questions that are prone to induce hallucination.
Specifically, we first sample $800$ documents from the \ourdata test set. For each document, we select the event which is \textit{non-factual} and have the most relation connections with other events in the document.\footnote{Having more relation connections suggests that the event involves more knowledge in the document, making it easier to construct knowledge-intensive questions about the event.}
We then utilize GPT-4 to generate yes-or-no questions that require complex reasoning along with the answers for each event based on its mentioned document. 
Then three experts manually review all the questions and answers, and eliminate questions without answers or not requiring reasoning, and correct erroneous answers, resulting in $450$ validated instances. 
An example of the data is shown in Table~\ref{tab:example_qa}. We evaluate two representative LLMs, LLAMA 3 and GPT-4. For adding event factuality, we adopt two settings: (1) Oracle setting, which adds the ground truth factuality. This setting allows controlled experiments to observe the efficacy of adding factuality. (2) Real-world setting, which adds the DMRoBERTa predicted factuality and aligns with real-world scenarios. More experimental details are placed in \cref{sec:append_hallu}.


\begin{table}[t]
    \centering
    \small
    \begin{tabular}{lcrr}
    \toprule
    \textbf{Setting} & \textbf{Factuality Info} & \textbf{LLAMA 3} & \textbf{GPT-4} \\
    \midrule
    Vanilla & \ding{55} & $77.6$ & $83.3$ \\
    Real-World & \ding{51} & $86.2$ & $94.4$ \\
    Oracle & \ding{51} & $88.9$ & $97.8$ \\
    \bottomrule
    \end{tabular}
    \caption{Accuracy (\%) on the constructed QA dataset. ``Vanilla'' denotes not adding factuality information.}
    \label{tab:mitigate_hallucination}
\end{table}

\paragraph{Experimental Results}
The results are shown in Table~\ref{tab:mitigate_hallucination}. We can observe that adding factuality information (Oracle setting) significantly improves the accuracy of LLMs, i.e., reducing the hallucination rate. Using the factuality automatically extracted using DMRoBERTa is also effective.
It offers a promising direction for research on reducing event-related hallucinations, namely by integrating additional factuality detection tools to explicitly include key information such as the factuality of events, triplets in the input, thereby mitigating model hallucinations. We encourage further exploration using \ourdata on this topic and to investigate more potential applications.

\section{Related Work}
\subsection{Event Factuality Detection Datasets}

FactBank~\citep{sauri2009factbank} is one of the earliest and widely-used EFD datasets. It is constructed based on TimeBank~\citep{timebank} and includes $5$ types of factuality. 
MEANTIME~\citep{minard2016meantime} re-annotated a subset of FactBank aimed at identifying the contextual factors influencing readers' veridicality judgments. To represent richer factuality, UW~\citep{lee2015event} and UDS-IH2~\citep{rudinger2018neural} adopted a continuous factuality value with a $[-3, 3]$ range. Recently, some studies introduced document-level EFD datasets, such as EB-DLEF~\citep{zhang2022evidence} and DLFE-v2~\citep{qian2022document}. \ourdata adopts $5$ discrete factuality categories to enhance annotation quality and reduce subjective bias, which we believe sufficiently represents factuality.

\subsection{Event Factuliaty Detection Methods}
Conventional event factuality detection methods primarily use neural-based models, mainly including developing novel network architectures~\citep{rudinger2018neural, qian2018event2, veyseh2019graph, cao2021uncertain, liu2022end} and designing new objectives~\citep{qian2018event, qian2019document, zhang2023code}. In the era of generative models, \citet{murzaku2023towards} transformed the event factuality detection task into a text generation form, utilizing FLAN-T5~\citep{chung2024scaling} for factuality detection. In this paper, we also evaluate LLMs and find that \ourdata poses significant challenges to existing methods. 

\section{Conclusion}
This paper introduces \ourdata, the largest and high-quality event factuality detection dataset. \ourdata comprehensively includes supporting evidence for factuality and event annotations from \maven. Experimental results demonstrate that \ourdata poses a significant challenge to EFD models and LLMs. We also find that using event factuality can help in mitigating event-related hallucinations in LLMs. We hope that \ourdata will facilitate research on the development and application of event factuality detection.
\section*{Limitations}
We discuss the limitations of this work here: 
(1) Language coverage. \ourdata only supports English, which may limit the widespread usage and application of our data. In the future, we will try to cover more languages and encourage community efforts for developing multilingual \ourdata.
(2) Annotation approach. Our annotation approach only saves approximately $15\%$ annotation cost. To ensure quality, we still employ substantial human annotation. However, this $15\%$ reduction means a saving of about $2,500$ USD. We encourage the community to develop more advanced automated annotation methods using \ourdata.
(3) LLM performance. We do not explore more prompting methods to enhance the performance of LLMs. We think this does not affect our experimental conclusions. LLMs typically underperform in specification-heavy tasks~\citep{peng2023specification} and require further efforts to improve their performance in EFD task.


\section*{Ethical Considerations}
We discuss ethical concerns here:
(1) \textbf{Intellectual property.} The \mavened dataset is released with CC BY-SA 4.0 license\footnote{\url{https://creativecommons.org/licenses/by-sa/4.0/}}. The \mavenarg and \mavenere are published with GPLv3\footnote{\url{https://www.gnu.org/licenses/gpl-3.0.html}} license. We strictly adhere to their licenses when using these data.
(2) \textbf{Intended use.} \ourdata is an event factuality detection dataset. Researchers and developers can use \ourdata to develop more advanced EFD methods and applications.
(3) \textbf{Potential risk control.} \ourdata is constructed from public data, which we believe has been well anonymized and desensitized. The data annotation process does not include any personal or sensitive information of the annotators. We believe \ourdata introduces no additional risks. We will not release the test set and instead use an online scoring platform following previous work~\citep{rajpurkar-etal-2016-squad,wang2020maven,wang2022maven,wang2023maven} to prevent potential cheating use and data contamination~\citep{xu2024benchmark}, thereby ensuring a fair evaluation.
(4) \textbf{AI assistance.} The writing of this paper employs ChatGPT to paraphrase some sentences.

\bibliography{custom}

\clearpage
\appendix

\section*{Appendices}
\section{Details on Data Construction}
\label{sec:app_data}

This section introduces details on human annotation of \ourdata, including details of annotation instruction (appendix~\ref{subsec:app_anno_inst}) and annotation coordination (appendix~\ref{subsec:app_anno_coor}).

\subsection{Annotation Instruction}
\label{subsec:app_anno_inst}
\begin{figure*}[t]
    \centering
    \includegraphics[width=0.9\linewidth]{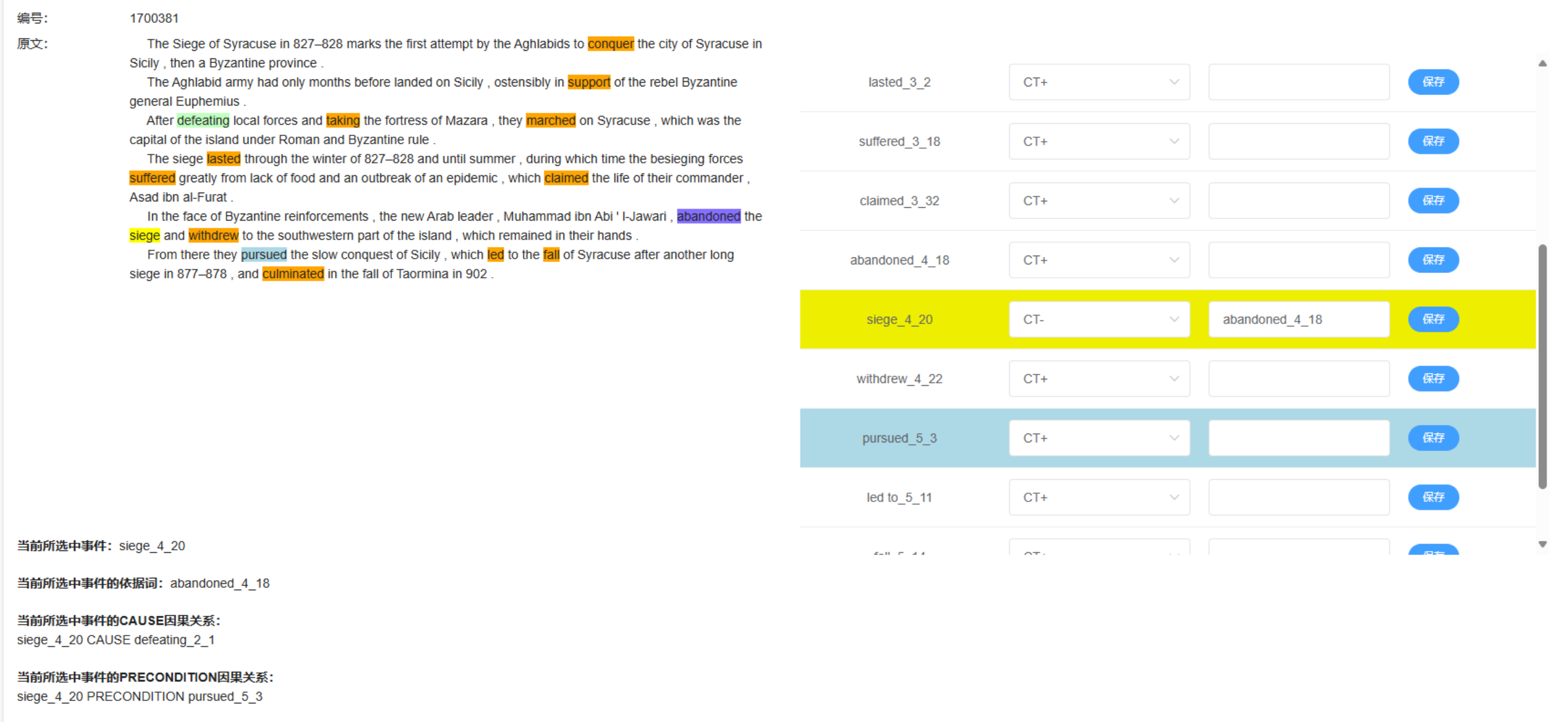}
    \caption{Screenshot for the annotation platform. The trigger word ``siege'' is selected for annotation, highlighted in yellow. Events related to it are highlighted in blue and green based on their relation type.}
    \label{fig:anno_plat}
\end{figure*}

Events pre-labeled as \textit{non-CT+} in the training set and events in the validation and test sets are manually annotated. During the annotation process, we incorporated heuristic rules derived from contextual information and event relations from \mavenere~\cite{wang2022maven} to guide the annotators. Some examples of the annotation rules can be seen in Table~\ref{tab:anno_rules}. For each factuality label, we provided specific examples and detailed explanations. This made it easier for the annotators to differentiate accurately. Additionally, we developed a new online annotation platform to support efficient and precise annotation, as shown in Figure~\ref{fig:anno_plat}.

\subsection{Annotation Coordination}
\label{subsec:app_anno_coor}

We engage annotators from a commercial data annotation company, comprising senior levels for annotation training and quality verification and others for data annotation. There are $47$ annotators in total, among whom $55\%$ are male and $45\%$ are female. All annotators receive fair compensation, with their salaries and workloads agreed upon in advance. Employment is contract-based and adheres to local regulations. The total cost for annotation, including both the factuality and supporting evidence, as well as the development of annotation platforms, amounts to approximately $14,000$ USD. We explained how the data would be used and obtained consent.

\section{EFD Experimental Details}
\label{sec:app_efd}
In this section, we introduce the implementation details regarding general details (appendix~\ref{subsec:gen_impl}) and task-specific details (appendix~\ref{subsec:task_impl}).

\begin{table}[t]
    \centering
    \small
    \begin{adjustbox}{max width=1\linewidth}{
    \begin{tabular}{l|c}
    \toprule
    \textbf{Model} & \textbf{Checkpoint / API}  \\
    \midrule
    BERT / DMBERT  &  \texttt{bert-large-uncased} \tablefootnote{\url{https://huggingface.co/google-bert/bert-large-uncased}} \\
    RoBERTa / DMRoBERTa & \texttt{roberta-large}\tablefootnote{\url{https://huggingface.co/FacebookAI/roberta-large}} \\
    GenEFD & \texttt{flan-t5-base}\tablefootnote{\url{https://huggingface.co/google/flan-t5-base}}\\
    LLAMA 3 & \texttt{Meta-Llama-3-8B-Instruct}\tablefootnote{\url{https://huggingface.co/meta-llama/Meta-Llama-3-8B-Instruct}} \\ 
    Mistral 7B & \texttt{Mistral-7B-Instruct-v0.2}\tablefootnote{\url{https://huggingface.co/mistralai/Mistral-7B-Instruct-v0.2}}\\
    GPT-3.5 & \texttt{gpt-3.5-turbo} \\
    GPT-4 & \texttt{gpt-4} \\
    \bottomrule
    \end{tabular}
    }
    \end{adjustbox}
    \caption{The correspondence between model and checkpoints or APIs.}
    \label{tab:ckpt}
\end{table}

\subsection{General Implementation Details}
\label{subsec:gen_impl}
For fine-tuned EFD models, we train the models on our train set with a learning rate of $1e-5$ and a batch size of $16$ over $10$ epochs based on their checkpoint from HuggingFace. Table~\ref{tab:ckpt} shows the correspondence between models and the checkpoints we used for training. We insert special tokens (<e> and </e>) around the trigger words of events in the text to indicate their positions, which are also used as the basis for dynamic multi-pooling for DMBERT and DMRoBERTa.

For large language models with in-context learning, we use the official OpenAI to evaluate GPT-3.5 and GPT-4, with the decoding sampling temperature set to 0, and other parameters kept as default. We utilize the checkpoints from HuggingFace to evaluate LLAMA 3 and Mistral 7B. The checkpoints and API we used are also shown in Table~\ref{tab:ckpt}.

All experiments are performed in a single run. We conduct experiments on Nvidia GeForce RTX $3090$ GPUs, totaling approximately $200$ GPU hours. For GPT-3.5 and GPT-4, we spend about $300$ USD in total. 

\subsection{Task Specific Details}
\label{subsec:task_impl}

For the event factuality detection task, we conduct sentence-level training and testing, each data item is a sentence with its marked events. We conduct GenEFD experiments with the prefix ``Event factuality prediction'' for each data item. The prompt used in the in-context learning experiments of large language models is listed in Table~\ref{tab:prompt_efd}.

For the supporting evidence prediction task, we approach it as a token classification task. The input consists of a list of words in the sentence and we insert special tokens showing its factuality around the trigger word of the event. The output is a list of the same length, with each element indicating the type of the corresponding word. In the output, `O' represents other types, `B' signifies the beginning of supporting words and `I' indicates the interior of supporting words. This definition applies to the fine-tuned EFD models. For large language models, to enhance their understanding of the task, we provide prompts in addition to the input, as shown in Table~\ref{tab:prompt_swp}.

For the task interaction, we utilize different approaches depending on the models and the tasks. In terms of tasks, for event relations, we mark triggers that have causal relations to the event to be classified with different special tokens in the sentences, and then concatenate them. For event arguments, we arrange them in a Type, Entity key-value pair format. Regarding the models, for DMRoBERTa, we use the processed event relations and event arguments as the input of its encoder, using the average representation of the token sequence as their representation. These representations are then concatenated with the original representation for classification. For GenEFD, we directly concatenate the processed event relations and event arguments with the original input as the model's input. For large language models, we incorporate explanations of event relations and arguments into the original event factuality detection prompt, followed by the processed event relations and arguments. The newly added prompt parts is shown in Table~\ref{tab:prompt_tl}.

\begin{table*}[t]
    \centering
    \small
    \begin{tabular}{p{0.9\linewidth}}
    \toprule
    \textbf{Modality Rules}  \\
    1. Modality is labeled based on the context. Events with a possible (\textit{PS}) modality usually have obvious hint words in the context, such as words with a tentative meaning like ``try to'', ``seek to'', or words indicating possibility like ``may'', ``might''. \\
    2. The Factuality of an event needs to consider its narrative source. We regard the document itself as the standard source. If the narrative source of an event is an argument in the document and the narration includes the argument's subjectivity, the modality is possible (\textit{PS}). \\
    \midrule
    \textbf{Polarity Rules}  \\
    1. Polarity is labeled based on the context. Events with a negative (-) modality usually have obvious hint words in the context, such as negative cues like ``prevent'', ``can not''. \\
    \midrule
    \textbf{Relation Rules}  \\
    1. For the event set $\mathbb{B}$ containing all the events \textit{B} with the relation \textit{B CAUSES A}, if any event \textit{B} in $\mathbb{B}$ has the factuality of \textit{CT+}, then the factuality of event \textit{A} is \textit{CT+}. \\
    2. For the event set $\mathbb{B}$ containing all the events \textit{B} with the relation \textit{B PRECONDITIONS A} relationship, if the factuality of event \textit{A} is \textit{CT-}, then the factuality of any event \textit{B} in $\mathbb{B}$ is \textit{CT-}. \\
    \bottomrule
    \end{tabular}
    \caption{Some annotation rules for human annotation process. ``Modality Rules'', ``Polarity Rules'' and ``Relation Rules'' represent the rules for classifying modality, classifying polarity, and utilizing relations, respectively.}
    \label{tab:anno_rules}
\end{table*}

\begin{table*}[t]
    \centering
    \small
    \begin{tabular}{p{0.9\linewidth}}
    \toprule
    \textbf{INSTRUCTION: } \\
    You are an event factuality classifier. Please annotate the factuality of events within the text. The events are marked with '<e>' and '</e>'. Assign one of the following five labels to the event:\\ CT+: The event is certain to happen or have happened.\\ CT-: The event is certain not to happen or have happened.\\ PS+: The event is possible to happen or have happened.\\ PS-: The event is possible not to happen or have happened.\\ Uu: The event factuality is unknown.\\
    \midrule
    \textbf{RULES: } \\
    Here are some annotation Rules:\\ 
    1. Firstly, observe the narrative source of the event. If the event originates from roles within the document instead of the document itself, label it 'PS+' or 'PS-'.\\ 
    2. Then, look for contextual cues like words that suggest attempts, such as 'try to' or 'seek to', and words indicating possibility, such as 'may' or 'might'. These events will also be labeled 'PS+' or 'PS-'.\\
    3. Events with indicators in the context, such as words conveying negative cues like 'stop' or 'prevent', will also be labeled 'PS-' or 'CT-'. \\
    \midrule
    \textbf{INPUT: } \\
    Here is the text you need to generate the label for, please do not output other information other than the label.\\
    TEXT: President Herbert Hoover then <e>ordered</e> the Army to clear the marchers' campsite. \\
    LABEL: \\
    \bottomrule
    \end{tabular}
    \caption{An example of prompt for event factuality detection task. The RULES part is used for inference with chain-of-thought prompt. The INPUT part varies depending on the data item.}
    \label{tab:prompt_efd}
\end{table*}

\begin{table*}[t]
    \centering
    \small
    \begin{tabular}{p{0.9\linewidth}}
    \toprule
    \textbf{INSTRUCTION: } \\
    Task Description: \\
    You are given a list of tokens representing a sentence containing an event. The event is marked with `<{factuality}>' and `</{factuality}>', where `factuality' indicates the factuality of the event. The possible values for factuality are:\\
     - CT+ (certainly happened) \\
     - CT- (certainly did not happen) \\
     - PS+ (possibly happened) \\
     - PS- (possibly did not happen) \\
    \\
    Your task is to generate an output list. Each element in the output list should correspond to an element in the token list. Use the following tags: \\
     - `O' for tokens that are not part of an evidential basis. \\
     - `B' for the beginning of an evidential basis. \\
     - `I' for the inside of an evidential basis. \\
     \\
     Carefully analyze the input sentence and identify the event marked by `<{factuality}>' and `</{factuality}>'. Identify the evidential basis words that support the factuality of the event. Generate the output list with `O', `B', and `I' tags according to the given rules. \\
    Ensure your output matches the format and corresponds accurately to the input token list. \\
    \midrule
    \textbf{EXAMPLE: } \\
    For example: \\
    Input: [`Webster', `s', `confession', `did', `not', `<CT->', `match', `</CT->', `the', `forensic', `evidence', `.'] \\
    Output: [`O', `O', `O', `B', `I', `O', `O', `O', `O', `O', `O', `O'] \\
    In this case, the event is ``match'', its factuality is CT-, and the evidential basis is ``did not''. \\
    \midrule
    \textbf{INPUT: } \\
    Input: [`The', `driver', `applied', `the', `brakes', `and', `reversed', `the', `engine', `,', `but', `was', `unable', `to', `<CT->', `stop', `</CT->', `in', `time', `.'] \\
    Output: \\
    \bottomrule
    \end{tabular}
    \caption{An example of the prompt for supporting evidence prediction task. The INPUT part varies depending on the data item.}
    \label{tab:prompt_swp}
\end{table*}

\begin{table*}[t]
\centering
\small
\begin{tabular}{p{0.9\linewidth}}
    \toprule
    \textbf{EXPLANATION: } \\
    In addition to the text, the event is accompanied by CAUSE relations, which is the <c>event</c> that causes the <e>event</e>, PRECONDITION relations, which is the <p>event</p> that must happen before the <e>event</e>. You can use these relations to help you determine the factuality of the event.\\
    \\
    Argument information is also provided for the event. The arguments are the entities that are involved in the event. You can use the arguments to help you determine the factuality of the event.\\
    \midrule 
    \textbf{RELATIONS: } \\
    Cause Relations: \\
    Most of them were <c>car bombs</c> and most targeted infrastructure, especially the transport network.\\
    At least twenty bombs <c>exploded</c> in the space of eighty minutes, most within a half hour period. \\
    Precondition Relations: \\
    The bombings were partly a response to the breakdown of <p>talks</p> between the IRA and the British government. \\
    \midrule 
    \textbf{ARGUMENTS: } \\
    Arguments:TYPE: Agent; ENTITY: IRA. TYPE: Location; ENTITY: Belfast. \\
    \bottomrule
\end{tabular}
\caption{An example of additional prompt for task interaction compared to event factuality detection task. The RELATIONS part and ARGUMENTS vary depending on the data item.}
\label{tab:prompt_tl}
\end{table*}
\section{Mitigating Hallucination}
\label{sec:append_hallu}

We employed GPT-4 (\texttt{gpt-4o-2024-05-13}) to construct the QA dataset based on \ourdata.
The construction process can be divided into two stages,  and Table~\ref{tab:prompt_constructQA} displays the prompt templates used in each stage.
Moreover, Table~\ref{tab:prompt_wo_wF} presents the prompt information corresponding to three test configurations.

\begin{table*}[ht]
    \centering
    \small
    \begin{tabular}{p{0.9\linewidth}}
    \toprule
    \textbf{STEP1: Constructing Reasoning Chain } \\
    Please generate an incorrect reasoning chain containing the "\{trigger\}" event (marked with <e> and </e>) based on the following document. \\
    Whenever possible, use the "\{trigger\}" event as the start or middle node event of the reasoning chain, rather than as the conclusion event.
    Note that the "{trigger}" event may not occur in the document, but in the reasoning chain, the "\{trigger\}" event must occur for the reasoning to be valid. \\
    This chain of reasoning should try not to include events not mentioned in the document.  \\
    Please give the chain of reasoning in numerical order and the reasoning chain within 6 steps. \\
    Document: \{document\}  \\
    Reasoning Chain:\\
    \midrule
    \textbf{STEP2: Constructing Question } \\
    Please give a question based on the above chain of reasoning. It should not be too simple or too difficult. \\
    The question should satisfy the following conditions: If the question is answered based on the above chain of reasoning, the answer will be Yes. However, if the question is answered based on the fact that "\{trigger\}" in the chain of reasoning does not necessarily occur, the answer will be No. Please directly output the questions that meet the requirements and do not output others. \\
    \bottomrule
    \end{tabular}
    \caption{Prompt template for constructing QA dataset based on \ourdata. The input and output of STEP1 are attached to the input of STEP2 as history.  In real application, \{trigger\} and \{document\} are filled with their corresponding input entries.}
    \label{tab:prompt_constructQA}
\end{table*}

\begin{table*}[ht]
    \centering
    \small
    \begin{tabular}{p{0.9\linewidth}}
    \toprule
    \textbf{Vanilla Setting} \\
    Please answer the questions according to the document below. Please answer Yes or No directly and do not enter other words. \\
    Document: \{document\} \\
    Question: \{question\} \\
    Answer: \\
    \midrule
    \textbf{Real-World and Oracle Setting} \\
    Please answer the questions according to the document below. \\
    Please carefully distinguish which events actually occurred in the document and which events are just possible events. Answer the questions based on what exactly happened in the document. \\
    Please answer Yes or No directly and do not enter other words. \\
    Document: \{document\} \\
    Question: \{question\} \\
    Note that the "\{trigger\}" event in the above document is not an exact occurrence, but a \{factuality\}. \\
    Answer:  \\
    \bottomrule
    \end{tabular}
    \caption{Prompt template for Vanilla, Real-World, and Oracle Setting in \cref{sec:hallucination}. In application, \{document\}, \{question\}, \{trigger\}, and \{factuality\} are populated with their respective inputs. Depending on the trigger's event factuality, \{factuality\} is assigned accordingly: "probable occurrence" for \textit{PS+}, "probable non-occurrence" for \textit{PS-}, and "definite non-occurrence" for \textit{CT-}.}
    \label{tab:prompt_wo_wF}
\end{table*}

\end{document}